\documentclass[letterpaper]{article}
\usepackage{aaai}
\usepackage{times}
\usepackage[]{algorithm2e}
\usepackage{helvet}
\usepackage{float}
\usepackage{courier}
\usepackage{graphicx}
\usepackage{subcaption}
\usepackage{array}
\usepackage{multirow}
\begin{document}
%
\title{Localized Flood Detection With Minimal Labeled Social Media Data Using Transfer Learning}


\author{Neha Singh, Nirmalya Roy, Aryya Gangopadhyay\\
Department of Information Systems\\
University of Maryland Baltimore County\\
Maryland, USA 21250\\
}

\maketitle
\begin{abstract}
\begin{quote}
Social media generates an enormous amount of data on a daily basis but it is very challenging to effectively utilize the data without annotating or labeling it according to the target application. We investigate the problem of localized flood detection using the social sensing model (Twitter) in order to provide an efficient, reliable and accurate flood text classification model with minimal labeled data. This study is important since it can immensely help in providing the flood-related updates and notifications to the city officials for emergency decision making, rescue operations, and early warnings, etc. We propose to perform the text classification using the inductive transfer learning method i.e pre-trained language model ULMFiT and fine-tune it in order to effectively classify the flood-related feeds in any new location. Finally, we show that using very little new labeled data in the target domain we can successfully build an efficient and high performing model for flood detection and analysis with human-generated facts and observations from Twitter.
\end{quote}
\end{abstract}

\section{Introduction}

There are various forms of a natural disaster such as flood, earthquake, volcano eruptions, storms, etc. but the flood is one of the lethal and prominent forms of natural disaster according to World Meteorological Organization (WMO) for most of the countries. National Weather Services (NWS) reported 28,826 flash floods events in the United States from October 2007 to October 2015 which resulted in 278 live loss and million-dollar worth crop and property damage \cite{gourley2017flash}. Monitoring and detecting floods in advance and proactively working towards saving peoples live and minimizing damage at the same time is amongst one of the most important tasks nowadays. In recent times, humans are extremely active on social media such as Twitter, Facebook, Youtube, Flickr, Instagram, etc. People use these platform extensively to share crucial information via message, photos and videos in real-time on social media for their interaction and information dissemination on every topic and acts as an active human sensor. It has been observed in the past few years via several case studies that social media also contributes significantly and being used extensively for crisis-related feeds \cite{yu2018big} and extremely helpful in situation awareness towards crisis management \cite{kim2018social,palen2018social,imran2015processing}. Emergency first responders agency, humanitarian organizations, city authorities and other end users are always looking for the right amount and content that would be helpful in the crisis scenarios but generally, social media provides an overwhelming amount of unlabeled data and it is very crucial to filter out the right kind of information using text classification. The advances in Artificial Intelligence (AI) which includes machine learning and Natural Language Processing (NLP) methods can track and focus on humanitarian relief process and extract meaningful insights from the huge amount of social media data generated regularly in a timely manner.

One of the major challenge while building a reliable and high accuracy model, it needs a huge amount of labeled data in order to be evaluated properly and achieve higher accuracy. Some of the platforms which uses crowdsourcing services and manually observe the data to label the disaster-related information such as CrisisLex\cite{olteanu2014crisislex}, CrisisNLP\cite{imran2016twitter}, CrisisMMD\cite{alam2018crisismmd}, AIDR\cite{imran2014aidr} etc. with already labeled data and pre-trained models, we can efficiently utilize the learned knowledge for the new target domain.

\begin{figure}[htbp]
\centerline{\includegraphics[width=9cm]{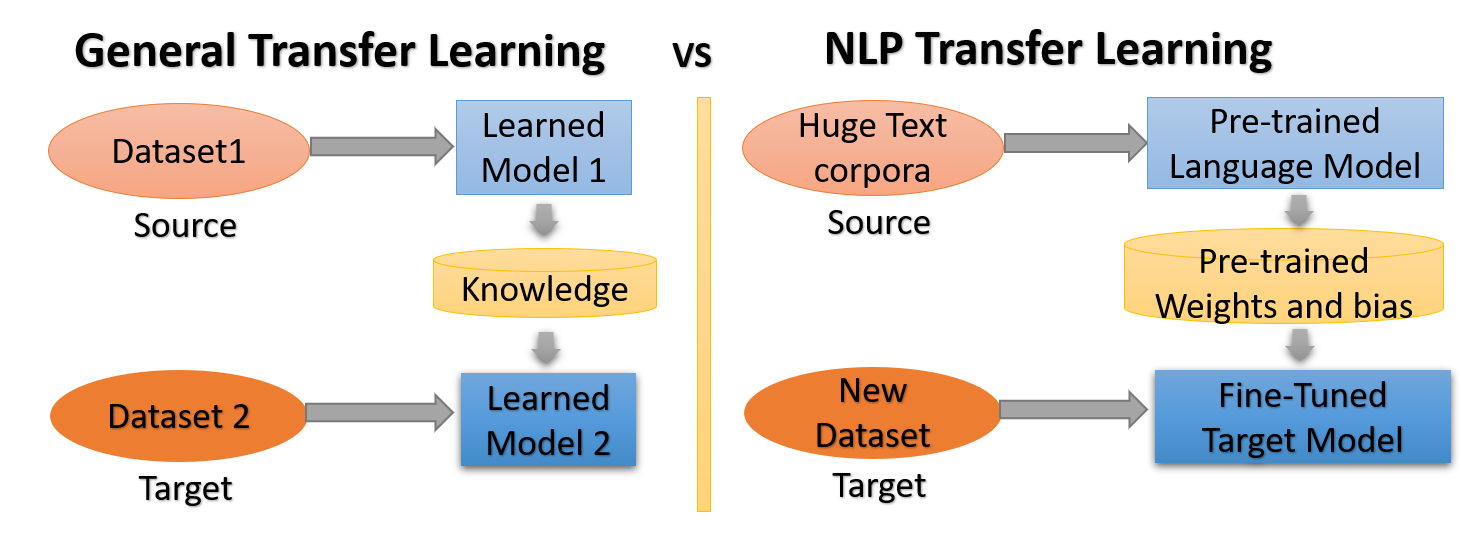}}
\caption{General Transfer Learning VS NLP Transfer Learning }
  \label{fig:comparisionTL}
\end{figure}

In general, to make a good predictive model we need a huge amount of labeled data with specific domain to train that provide accurate, reliable results for the new domain. Transfer learning models efficiently leverage the existing knowledge and perform effectively the intended task by adapting to the new domain. 
In Figure \ref{fig:comparisionTL} shows the comparison of general transfer learning and NLP transfer learning. Transfer learning learns from the source data model and applies the gained knowledge from the source domain to the target domain that requires relatively less labeled data. Social media growth in last decade and availability of existing disaster-related data sources labeled by crowdsourcing platforms provide an opportunity to utilize this data and build a learning model which learns the domain knowledge and transfer the learned knowledge to classify new data with higher accuracy and confidence automatically. 
This can effectively solve some of the important problems in disaster management such as flood detection, executing rescue operations, sending feedback and contextual warnings to authorities, improved situation awareness, etc. Transfer learning contains various type of knowledge sharing such as inductive, transductive depending on the source and target domain data distribution and source/target task relatedness \cite{pan2009survey}. Figure \ref{fig:comparisionTL} shows basic transfer Learning concept in NLP is slightly different than the general transfer learning. In general transfer learning, we have source domain and target domain, the model build and learned from the source domain data is used to transfer the knowledge to the target domain task model. Whereas, in NLP the source domain is the general understanding of the text learned from not only one domain but from a giant corpus of text, build a language model known as a pre-trained language model. These pre-trained language models are further used for different downstream task such as text classification, spam detection, question answering, etc. We are using here the inductive transfer learning where we have a pre-trained model as source task and improve the performance of the target task (flood tweet classification). We present in this study that using a pre-trained model and very few labeled flood tweets we can achieve great accuracy effectively in no time. 

The main contributions of this work are as follows:  
\begin{itemize}
    \item We propose to use the inductive transfer learning method and adapt the ULMFiT Pre-train model for text classification.
    \item We fine-tune the target model parameters by knowledge obtained from the source domain for quick and efficient flood tweet classification.
    \item We show that ULMFiT method needs a very small amount of labeled data (5\%) to achieve high accuracy and performance.
    \item This study demonstrates that this model can be applied in real-time flood detection and information extraction with very small training data for new application domain.
\end{itemize}


\section{Related Work}
Growing active user base on social media and has been created a great opportunity for extracting crucial information in real-time for various events and topics. Social media is being vigorously used as the communication channel in the time of any crisis or any natural disaster in order to convey the actionable information to the emergency responders to help them by more situational awareness context so that they make a better decision for rescue operations, sending alerts, reaching out people right on time. There have been numerous works proposed related to crisis management using social media content which is discussed in the following section. 

\textbf{Social media for crisis management}
Mainly in the analysis of social media content related to crisis situations data type such as images, geolocation, videos, text, etc. but most of the focus of these work has been images and geolocation towards crisis management \cite{kim2018social,palen2018social,imran2015processing,singh2019analyzing}. Processing social media content is itself a huge challenge and comes with great challenges as well such as information processing, cleaning, filtering, summarizing, extracting, etc. There has been some progress lately in developing methods to extract meaningful information during a crisis for better situation awareness and better decision making \cite{keim2011emergent}. The text domain of the social media data has not been exploited to its fullest and it is generally the most valuable and available data on social media. Text processing can provide great amount of details which can be useful for situation awareness and help towards extracting actionable insights.
Identifying relevant text data would eventually result in major event detection which is difficult to correctly track in a short amount of time and fast processing is needed in these scenarios. \cite{keim2011emergent,singh2019analyzing}. 

\textbf{Domain adaptation for crisis management}
Transfer learning is very popular and active research area of machine learning. This learning method is known for learning the domain knowledge while solving the task and transfer its knowledge from one domain (source) to another domain (target) to solve the task in the new domain. We need to know these basic things while applying transfer learning (1). What needs to be transferred? (2). When to transfer the learned knowledge? (3). How to transfer knowledge? There are few basic transfer learning algorithm principles that include few simple steps as follows: (i) it aims to minimize the error measure by reweighting the source label sample such that it appears as a target. (ii) Adapt the methods iteratively and label target example using these common steps (a) model learned from labeled example, (b) labels some target example (c) New model learns from new labels \cite{kaboli2017review,do2006transfer}. Transfer learning has been explored and applied in various classification problems for high quality and reliable results with less labeled data in the target domain. It has also been used for feature selection, pedestrian detection, improving visual tracking and subtractive bias removal in medial domain\cite{kaboli2017review}. Some of the other example where transfer learning have been used are text classification\cite{do2006transfer}, sentiment classification \cite{blitzer2007biographies,tan2009adapting}, domain adaptation \cite{li2015twitter}, object classification\cite{bergamo2010exploiting}.

\section{Data Collection and Processing}
In this section, we explain about our data collection and cleaning process of the data followed by some data visualization for better understanding of the data. The text data are decidedly very crucial and if leveraged carefully in time, it can assist in various emergency response services. It could greatly benefit the authorities in their decision-making process, rescue operation, increase situational awareness and early warnings. We are using Twitter data since it is one of the widely used social media platform in recent times.\\
\textbf{Data Collection:} We are using the disaster data from \cite{olteanu2014crisislex}. It contains various dataset including the \textbf{CrisiLexT6} dataset which contains six crisis events related to English tweets in 2012 and 2013, labeled by relatedness (on-topic and off-topic) of respective crisis. Each crisis event tweets contain almost 10,000 labeled tweets but we are only focused on flood-related tweets thus, we experimented with only two flood event i.e. \textit{Queensland flood in Queensland, Australia} and \textit{Alberta flood in Alberta, Canada} and relabeled all on-topic tweets as Related and Off-topic as Unrelated for implicit class labels understanding in this case. The data collection process and duration of CrisisLex data is described in \cite{olteanu2014crisislex} details. \\
\begin{table}[htbp]
\centering
\begin{tabular}{ |c|c| } 
\hline
\textbf{Class Label} &\textbf{Tweets Counts} \\ 
\hline
Related & 5414\\ \hline
Unrelated  & 4619 \\ \hline
\end{tabular}
\caption{Data Distribution}
\label{table:classlabel}
\end{table}

\begin{figure}[htpb]
    \centering
    \begin{subfigure}[t]{0.2\textwidth}
        \centering
        \includegraphics[height=1.2in]{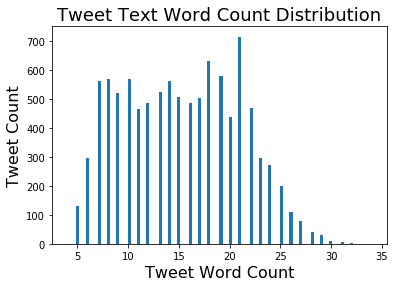}
        \caption{Word Distribution}
        \label{fig:Tword_dist}
    \end{subfigure}
    ~~~~~~~~
    \begin{subfigure}[t]{0.2\textwidth}
        \centering
        \includegraphics[height=1.2in]{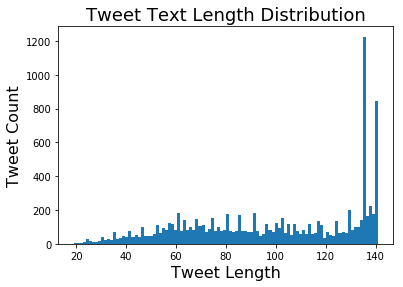}
        \caption{Tweet Length Distribution}
        \label{fig:Tlen}
    \end{subfigure}
\caption{Tweet Distribution}
\end{figure}
\textbf{Data cleaning:} The tweets, in general, are very noisy and we need to clean the tweets in order to use them for efficient model building. We removed the stop words, numerical, special symbols and characters, punctuation, white space, random alphabets, and URLs, etc. We also transform all the tweets into lower case alphabet to normalize it and remove the redundancy in the data. After cleaning the tweets we performed some data visualization next for better data insights.    \\
\begin{figure}[htbp]
    \centering
    \begin{subfigure}[t]{0.2\textwidth}
        \centering
        \includegraphics[height=1.7in]{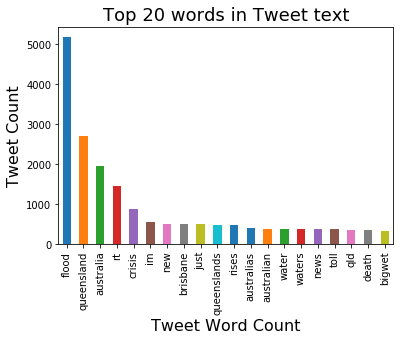}
        \caption{Frequent Words}
        \label{fig:count}
    \end{subfigure}%
  ~~~~~~~~~~~~~~~
    \begin{subfigure}[t]{0.2\textwidth}
        \centering
        \includegraphics[height=1.7in]{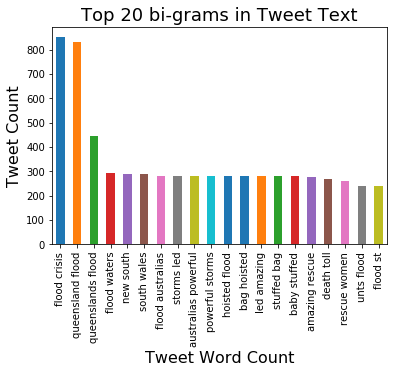}
        \caption{Bigram}
        \label{fig:bigram}
    \end{subfigure}
    ~\
    \begin{subfigure}[t]{0.2\textwidth}
        \centering
        \includegraphics[height=1.8in]{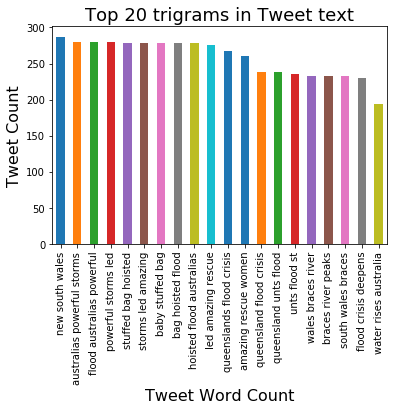}
        \caption{Trigram}
        \label{fig:trigram}
    \end{subfigure}
    \caption{Tweet Data Visualization}
\end{figure}

\textbf{Data Visualization:} Our focus here is to understand the basic characteristics of tweets and demonstrate the power of transfer learning method in this application. Although both of the datasets are similar in distribution thus, we have selected Queensland flood dataset for elaboration. Table \ref{table:classlabel} shows the fairly equal class distribution in Queensland flood tweets with 5414 related flood tweet and 4619 unrelated flood tweets. Figure \ref{fig:Tword_dist} shows the number of words in a tweet which ranges from 5 words up to 30 words in a single tweet. Figure \ref{fig:Tlen} shows the tweet length distribution contains from 30 characters up to 140 characters in a tweet. Figure \ref{fig:count}, \ref{fig:bigram}, \ref{fig:trigram} shows the top 20 most frequent words, bi-gram and tri-gram respectively of the tweet dataset. By visual inspection of these most frequent words, bigram and trigram, we have a general understanding of the major topics and themes in the data. Tweets characteristics are generally similar in most of the cases so it is highly probable that it can be effectively applied for other scenarios or new location as well. 

\section{Methodology}

It is well known that numerous state-of-the-art models in NLP require huge data to be trained on from scratch to achieve reasonable results. These models take paramount of memory and immensely time-consuming. NLP researchers have been looking into various successful methods/models in computer vision (CV) and to attain similar success in NLP. A major breakthrough in CV was transferring knowledge obtained from pre-trained models on ImageNet \cite{krizhevsky2012imagenet} as a source task to target tasks for efficient results.
There has been a huge advancement in the area of transfer learning in NLP due to the introduction of the pre-trained language models such as ULMFIT\cite{howard2018universal}, ELMO \cite{peters2018deep},GLUE \cite{wang2018glue}, BERT \cite{devlin2018bert}, Attention-net \cite{vaswani2017attention}, XL-Net \cite{yang2019xlnet} and many more to come etc.
These pre-trained models have acquired state-of-the-art performance for many NLP task since they use a huge amount of training data for language understanding as their source models and fine-tune the model to achieve the high accuracy in the target task. We are using ULMFiT in this study since it has been shown significant performance for target domain classification task with minimal labelled data along with less training time with reasonable hardware requirement. Whereas, other models such as BERT, XL-Net etc. are much bigger and complex that need large training time and higher hardware architecture.  

\subsection{Universal Language Model Fine-tuning (ULMFiT)}
This method ULMFiT \cite{howard2018universal} was introduced by Howard and Ruder which can effectively be applied as a transfer learning method for various NLP task. In inductive transfer learning the source task (Language model) is generally different than the target task (Flood detection) and requires labeled data in the target domain.
ULMFiT is very suitable for efficient and text classification \cite{howard2018universal} is a pre-trained model. This model significantly outperformed in text classification, reducing error by 18-24\% on various datasets and achieving accuracy with very small labeled data. Some of the examples where researchers have used ULMFiT to solve a specific problem using power of transfer learning are \cite{hepburn2018universal,rother2018ulmfit}.
Although, ULMFiT has the capability to handle any type of classification task such as topic classification, question classification, etc. but we are specifically targeting the flood-related tweet classification. 

\subsection{ULMFiT adaptation for Flood Tweet Classification}

\begin{figure}[htpb]
\centerline{\includegraphics[width=9cm]{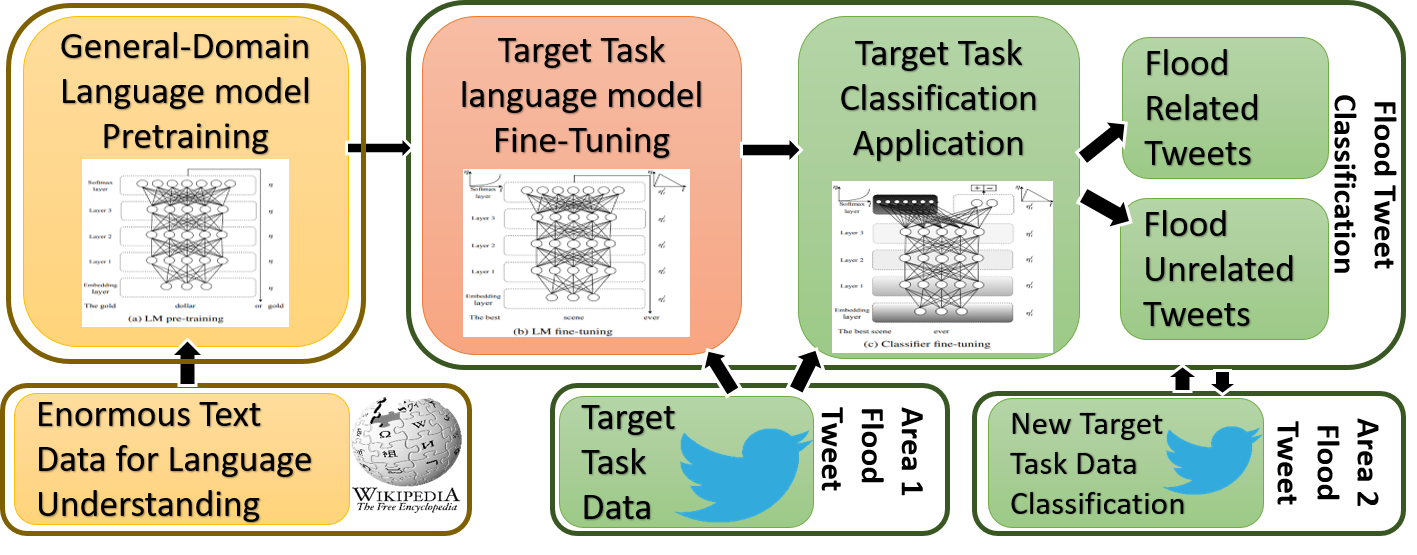}}
\caption{Flood Text Classification Framework adapted from \cite{howard2018universal}}
  \label{fig:framework}
\end{figure}

Text classification in any new area generally suffers from no or very little labeled data to work with initially. Inductive transfer learning addressees this very same challenge and ULMFiT method is primarily based on this concept.
We have used the pre-trained language model ULMFiT to do the classification for the target task and classify the related and unrelated flood tweets coming from different location social media (Twitter). As shown in Figure \ref{fig:framework} our overall framework adapted from \cite{howard2018universal} to do the flood tweet classification.
As shown in Figure \ref{fig:framework} we are using the ULMFIiT architecture to solve the flood tweet classification problem. The source domain here is trained on the paramount of text data corpus from WikiText-103 dataset which contains 103 million words, 400 dimensional embedding size, 3 layers neural network architecture (AWD-LSTM) and 1150 hidden activations per layer that creates a general domain language model for \textbf{general domain LM pretraining} to predict the next word in the sequence, learns general features of the language. 
AWD-LSTM \cite{merity2017regularizing} is a regular LSTM, used for the Language Modeling with various regularization and optimization techniques that produce state-of-the-art results. Next step is \textbf{Target Task LM Fine-Tuning} which entertain the transfer learning idea by gaining the knowledge from the previous step and utilize it in the target task. Here the target task is flood tweet detection which has different data distribution and features so the general model fine-tunes according to the target task and adapt to the new domain (target) by learning the target task-specific features of the language. It is done using discriminative fine-tuning and slanted triangular learning rates for fine-tuning the LM. Finally, \textbf{Target Task Classifier} provide classification results as the probability distribution over flood class labels (related and unrelated) which is a very critical part of transfer learning method. it needs to be very balanced (not too slow or fast fine-tuned) using the gradual unfreezing for fine-tuning the classifier. We used some of the same hyperparameters for this task.

\section{Experimental Results and Discussion}
In this section, we will discuss our experimental results of the text classification.
As described above in the methodology section that our source domain model comes from the ULMFiT and the target domain data is Queensland flood data which has almost 10,000 tweets labeled as flood \textbf{Related} and \textbf{Unrelated}. The pre-train ULMFiT model uses the AWD-LSTM language model with embedding size of 400, 3 layers, 1150 hidden activations per layer with a batch size of 70 and a back propagation through time (BPTT) \cite{howard2018universal}. Dropout here has been used as 0.7 to language model learner and 0.7 to text classifier learner. A base learning rate of 0.01 for LM fine-tuning and multiple values ranging from 0.00001 to 0.1 of learning rate have been used for target classifier fine-tuning for various instances. We have used gradual unfreezing of the model layers in this case to avoid the risk of catastrophic forgetting. It starts fine-tuning of the last layer (minimal general knowledge) to the next lower layer on wards in every iterations to attain the highest performance of the model.

\begin{figure}[htpb]
    \centering
    \begin{subfigure}[t]{0.2\textwidth}
        \centering
        \includegraphics[height=1.2in]{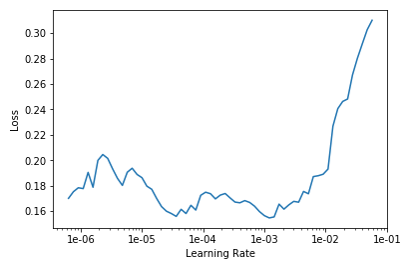}
        \caption{learning rate}
        \label{fig:lr}
    \end{subfigure}
    ~~~~~~~~
    \begin{subfigure}[t]{0.2\textwidth}
        \centering
        \includegraphics[height=1.2in]{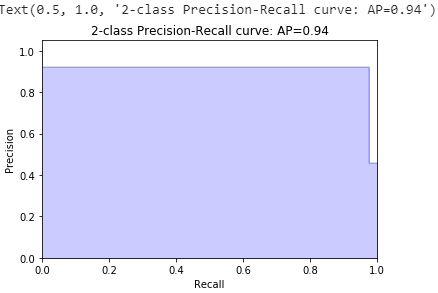}
        \caption{Precision-Recall curve}
        \label{fig:eval}
    \end{subfigure}
\caption{Data Evaluation Metrics for Queensland Data}
\end{figure}

We have used the following hardware for the experimentation:
Windows 10 Education desktop consisting of intel core i-7 processor and 16GB RAM. We have used python 3.6 and Google colab notebook to execute our model and obtained the results discussed below:  
The train and test data have divided into 70-30 ratio and we got these results as shown in Table \ref{table:accuracy} for the individual dataset and the combination of both. The pre-trained network was already trained and we used the target data Queensland flood which provided 96\% accuracy with 0.118 Test loss in only 11 seconds provided we used only 70\% of training labeled data. The second target data is Alberta flood with the same configuration of train-test split which provided 95\% accuracy with 0.118 Test loss in just 19 seconds. As we can see it takes very less time to work with 20,000 of tweets (combined) and at times of emergency it can handle a huge amount of unlabeled data to classify into meaningful categories in minutes.
\begin{table}[htpb]
\centering
\begin{tabular}{ |c|p{0.7cm}|p{0.7cm}|c|c| } 
\hline
\textbf{Target Data} & \textbf{Train \newline Loss} & \textbf{Test \newline Loss} & \textbf{Accuracy} & \textbf{Time(sec)} \\ 
\hline
Queensland & 0.162  & 0.118 & 0.960 & 00:11\\ \hline
Alberta  & 0.193 & 0.176 & 0.953 & 00:19 \\ \hline
Combined & 0.200 & 0.136 & 0.957 &00:19 \\\hline
\end{tabular}
\caption{Classification Accuracy Comparison}
\label{table:accuracy}
\end{table}
Here, Our focus is localized flood detection thus we are not merging multiple datasets, we will leave the combination for our future work and staying with one Queensland flood data and explore that in details. 
As it can be seen in Table~\ref{table:queens} that event with the 5\% of data which is only 500 labeled tweets as target labeled data the model can adapt and fine-tuned the classification model wit 95\% accuracy. This model is very efficient and effective when we have a time-sensitive application and instead of training a model from scratch with huge data we can use the pre-trained model and successfully applies to the target domain application. The Table~\ref{table:queens} also depicts that even with the very small labeled training data the model was able to achieve the accuracy almost equivalent to the 80\% of the training data.
There is generally a direct relation which says the more training data is the better but here increased labeled data the accuracy did not contribute significantly towards the accuracy improvement. 
\begin{table}[htpb]
\centering
\begin{tabular}{|p{1.2cm}|p{1.2cm}|p{1.1cm}|p{0.8cm}|p{.6cm}|p{1.2cm}|}\hline
\textbf{Labeled Data\% Target} & \textbf{Class Label} & \textbf{Precision} &\textbf{Recall} & \textbf{F1-score} &\textbf{Accuracy}\\ \hline
 5\% & Related \newline Unrelated &0.97\newline 0.93  & 0.94 \newline 0.97  &0.96 \newline0.95  & 0.95 \\ \hline
 10\% & Related \newline Unrelated & 0.98 \newline 0.94 & 0.95 \newline 0.97&0.96\newline0.96 & 0.96 \\ \hline
 20\% & Related \newline Unrelated & 0.97\newline 0.95 & 0.95\newline 0.97 &0.96\newline0.96 & 0.96   \\ \hline
 50\% &  Related \newline Unrelated& 0.98\newline0.93 &0.94\newline 0.98 &0.96\newline0.95&  0.96  \\ \hline
 80\% & Related \newline Unrelated &  
0.97\newline0.94 &0.95\newline 0.96 &0.96\newline
0.95 & 0.96    \\ \hline
\end{tabular}
\caption{Evaluation Metrics for Queensland Flood Data}
\label{table:queens}
\end{table}
There are some more measures for accessing the quality of the classification such as training/testing loss and average precision to avoid the bias in the accuracy. Thus, Figure \ref{fig:lr} shows the learning rate adjusting according to the target classifier model, showing with the specific learning rate it achieves the low amount of loss which is called as the slanted triangular Learning rate. Figure \ref{fig:eval} shows the Precision-Recall curve for a particular classification instance where the average Precision is 0.94. It shows that the overall quality of the classification is fairly good and does not favor one class over another.

As described above and based on the experimental results we can use a very low amount of labeled data and solved the localized flood disaster situation efficiently for any new location. We faced some limitations in this work and plan to include in our future work described in the next section.

\section{Limitation and Future Work}

We have been focused on a specific type of disaster (Flood) here and did not explore other disaster types since we wanted to capture specific kind of disaster characteristics and learn from it for another flood disaster. We plan to perform extensive experimentation with some other kind of disaster data as well in the future.

We have explored and experimented with the twitter dataset only so far because it is widely available and accessible for everyone but we would attempt to include different kinds of data sources such as other social media platforms, news feeds, blogs, text, images, etc. as well to make it a multimodel transfer learning approach in our future models.

There are other state-of-the-art pre-trained language model such as BERT, GPT-2, Transformer-XL, etc. for text classification available and we would want to compare this adaptation with other models as well for the most time effective models in the given situation. 

There can be many more application where multi-class classification including various classes such damage, rescue, buildings, transportation, medical, etc. can be labeled with a small amount in order to build a very efficient classification model. We also have the plan to formulate this a multi-class problem in order to deeply address the problems in disaster management.  

This opens up a new door for cyber-physical-social systems that would rely on social media feeds coming from human sensors along with wireless physical/environmental sensors in tandem for various applications to create another layer of smart sensors that can achieve the high quality, more reliable and fault-tolerant system.

\section{Conclusion}
As we are aware of the calamity due to flood/ flash flood situation which needs close monitoring and detail attention. With the exponential growth in social media users, there is an ample amount of data which can be extremely useful in flood detection. Transfer learning is very helpful in these applications where we need to train with general knowledge along with little target domain knowledge to attain a highly effective model. We have discovered that inductive transfer learning methods are very useful for social media flood detection data with minimal labeled data. We used Queensland Twitter data as one of the flood locations and used the pre-trained model ULMFiT to successfully classify with accuracy 95\% the flood-related tweets with only 5\% of labeled target samples under 10 seconds whereas in general, it takes thousands of labeled tweets and huge time to achieve the similar performance. The usage of pre-trained models with minimal space and time complexity, it can be a huge advantage to the time-sensitive application where we need to process millions of tweets efficiently and classify them accordingly with high performance without compromising on the accuracy.

\section{Acknowledgment}

This research is funded by the National Science Foundation (NSF) grant number 1640625. I would like to thank my mentor and advisor Dr. Nirmalya Roy for their motivation, support, and feedback for my research. I am grateful for Dr. Aryya Gangopadhyay (co-advisor) for the discussion and continuous encouragement towards my work.

\footnotesize
\bibliographystyle{aaai}
\bibliography{AAAI.bib}

\end{document}